\documentclass{article}
\usepackage[preprint]{neurips_2024}
\makeatletter
\renewcommand{\@noticestring}{Preprint.}
\makeatother
\usepackage[utf8]{inputenc}
\usepackage[T1]{fontenc}
\usepackage{hyperref}
\usepackage{url}
\usepackage{booktabs}
\usepackage{amsmath}
\usepackage{amssymb}
\usepackage{graphicx}
\usepackage{xcolor}
\usepackage{float}
\usepackage{subcaption}

\title{Good Scores, Bad Data: A Metric for Multimodal Coherence}

\author{Vasundra Srinivasan\thanks{This work represents the author's independent research conducted outside the scope of any employment or contractual obligation. It is not sponsored by, endorsed by, or affiliated with the author's employer or any organization referenced herein. No proprietary or confidential information is disclosed; all observations derive solely from publicly available datasets and open-source tools.} \\
Author, \emph{Data Engineering for Multimodal AI} (O'Reilly) \\
Stanford School of Engineering, Graduate Certificate (in progress)
}

\begin{document}

\maketitle

\begin{abstract}
Multimodal AI systems fuse images, text, and structured annotations into unified representations, then evaluate the result by downstream task accuracy. However, high accuracy on research benchmarks does not translate to high confidence in production workflows. Accuracy conflates two independent properties: how well the model performs and how coherent the underlying data is. A model can achieve strong Visual Question Answering (VQA) scores on internally contradictory inputs simply by exploiting answer distributions.

We introduce the Multimodal Coherence Score (MCS), a metric that evaluates fusion quality independent of any downstream model. MCS decomposes coherence into four dimensions (identity, spatial, semantic, and decision) with learned weights optimized via Nelder-Mead to maximize Spearman correlation with downstream performance. We evaluate MCS on 1,000 Visual Genome images using DETR for object detection, CLIP for semantic alignment, and ViLT for VQA, then validate transfer on 150 COCO 2017 images with no retraining. Across three fusion architectures (naive concatenation, contract-enforced filtering, and CLIP-based re-ranking), MCS discriminates fusion quality with higher sensitivity than task accuracy alone (Spearman $\rho = 0.093$ vs.\ $0.071$). Controlled perturbation experiments on 200 images at three corruption rates confirm that each dimension responds independently to its failure mode. Zero cross-talk.

MCS is lightweight, requires no human annotation, and runs end-to-end in a single Colab session. It does not stop at \emph{something broke}. It tells you \emph{what broke}.
\end{abstract}

\section{Introduction}

\begin{figure}[t]
\centering
\includegraphics[width=\linewidth]{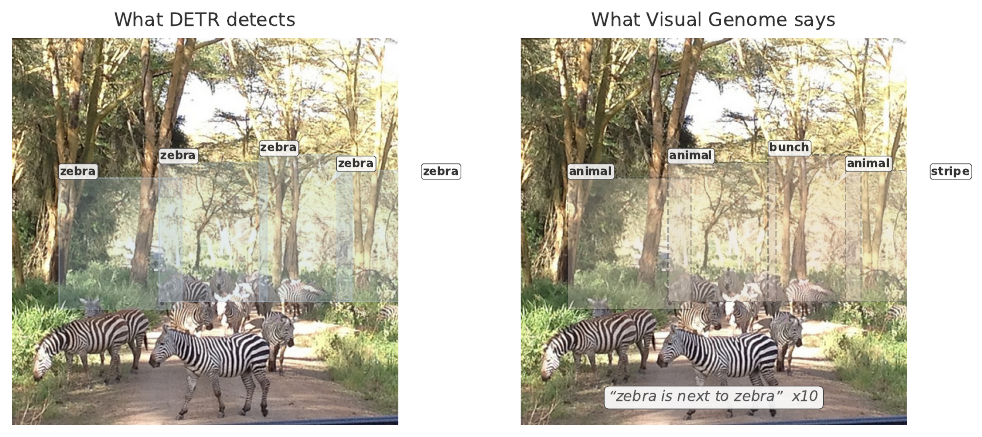}
\caption{Visual Genome image \#2320169. Left: what a DETR detector sees (``zebra'' for each animal). Right: what the dataset says (``animal,'' ``bunch,'' ``stripe,'' and ``zebra is next to zebra'' repeated ten times). A VQA (Visual Question Answering) system returns the correct answer. The benchmark score looks fine. The data is not.}
\label{fig:zebras}
\end{figure}

Consider Figure~\ref{fig:zebras}. Visual Genome image \#2320169 shows a herd of zebras crossing a dirt road. A DETR object detector identifies thirteen zebras. But the dataset annotations label most of them ``animal.'' One is labeled ``bunch.'' Individual stripes are listed as separate objects. The region descriptions repeat ``zebra is next to zebra'' ten times and ``animal with black and white stripes'' ten more. A Visual Question Answering (VQA) system is asked ``What kind of animal?'' and responds ``zebra.'' The answer is correct. The data behind it is not.

This matters beyond research benchmarks. The same class of incoherence, where structured annotations contradict each other within a single record, propagates directly into production systems that depend on multimodal fusion. A radiology pipeline where a scan's metadata disagrees with the accompanying report. An autonomous vehicle where LiDAR annotations and camera labels place the same object at different locations. An enterprise document processing workflow where extracted tables contradict the surrounding text. In each case, a downstream model may still produce a plausible output, but the underlying fusion is unreliable, and no existing metric flags it.

This is not an isolated quality issue. Visual Genome, the densest public multimodal dataset and the backbone of scene graph research, contains thousands of images where annotations disagree with each other \cite{krishna2017visual}. The field currently lacks a systematic tool for detecting this. Downstream metrics such as VQA accuracy, Recall@K, and BLEU \cite{vedantam2015cider} evaluate \emph{model} performance, not data quality. Pairwise alignment methods such as CLIP similarity \cite{radford2021learning} assess two modalities at a time. When a fusion event spans five modalities, pairwise checks do not capture the full picture. To date, no metric evaluates the coherence of the fusion itself.

We introduce the Multimodal Coherence Score (MCS), which decomposes fusion quality into four dimensions: identity coherence (do records refer to the same entities?), spatial coherence (do spatial relationships agree across modalities?), semantic coherence (is the content mutually consistent?), and decision coherence (does the fused representation support correct downstream answers?). A learned weighted composite produces a single score that is more predictive of downstream performance than any individual dimension. Perturbation experiments across three datasets confirm that each dimension responds independently to its targeted failure mode: swapping object annotations degrades identity coherence while all other dimensions hold steady.

The result is a metric that is efficient to compute, requires no human annotation, and provides actionable diagnostics. It does not stop at \emph{something broke}. It tells you \emph{what broke}, and which part of the pipeline to address first.

\section{Background}

\textbf{Multimodal evaluation.} CLIP alignment \cite{radford2021learning}, VQA accuracy, and captioning scores \cite{vedantam2015cider, banerjee2005meteor} measure how well a \emph{model} handles multimodal input. They do not assess whether the \emph{data} itself is coherent.

\textbf{Data quality.} Wang and Strong \cite{wang1996beyond} defined quality dimensions for structured databases. These frameworks assume a single-table setting and do not extend to cross-modal relationships.

\textbf{Scene graph evaluation.} SGDet, SGCls, and Recall@K \cite{krishna2017visual} evaluate predicted graphs against ground truth assumed to be clean. In Visual Genome, this assumption does not hold.

\textbf{The gap.} No existing framework measures cross-modal coherence as an intrinsic property of the fused representation, independent of downstream task performance. This paper addresses that gap.

\section{The Multimodal Coherence Framework}

\subsection{Setup}

Let $E = \{r_1, r_2, \ldots, r_n\}$ be a multimodal event: a set of records from $n$ modalities describing the same scene. In Visual Genome, one event per image contains: image pixels ($r_{\text{img}}$), object annotations ($r_{\text{obj}}$), relationship triplets ($r_{\text{rel}}$), region descriptions ($r_{\text{desc}}$), and QA pairs ($r_{\text{qa}}$).

Coherence means these records agree with each other. When they do not, the fusion is unreliable, even if a downstream model produces a correct output. Four dimensions capture the distinct ways fusion can fail.

\subsection{The Four Dimensions}

\textbf{Identity Coherence (IC).} Do all records refer to the same entities?

In image \#2320169, a detector identifies thirteen zebras. If the annotations also list ``zebra'' thirteen times, IC is high. Instead, they list ``animal,'' ``bunch,'' ``stripe.'' IC is low.

IC is the Jaccard similarity between detected and annotated entity sets:
\begin{equation}
\text{IC}(E) = \frac{|\text{entities}(r_{\text{obj}}) \cap \text{detected}(r_{\text{img}})|}{|\text{entities}(r_{\text{obj}}) \cup \text{detected}(r_{\text{img}})|}.
\end{equation}

We use DETR-ResNet-50 \cite{carion2020end} with a confidence threshold of 0.7 to define the detected entity set. A label normalization map aligns Visual Genome vocabulary to COCO categories (e.g., ``man,'' ``woman,'' ``boy'' $\to$ ``person'').

\textbf{Spatial Coherence (SpC).} Do spatial relationships match across modalities?

A region description places ``a zebra eating grass'' at bounding box coordinates $(6, 353, 51, 67)$. A detector localizes the same zebra nearby. SpC is high. But ten descriptions say ``zebra is next to zebra'' with overlapping boxes that fail to isolate individual animals. SpC is low.

SpC is the mean IoU between annotated and detected bounding boxes:
\begin{equation}
\text{SpC}(E) = \frac{1}{|R|} \sum_{i=1}^{|R|} \text{IoU}(\text{bbox}_i^{\text{anno}}, \text{region}_i^{\text{det}}).
\end{equation}

\textbf{Semantic Coherence (SC).} Is the content across modalities mutually consistent?

If descriptions said ``a boat at a marina'' for the zebra image, SC would be low. Even the actual annotations (``animal with black and white stripes'' repeated ten times, with no mention of the road or trees) produce lower SC than a single well-written caption.

SC uses CLIP ViT-B/32 embeddings \cite{radford2021learning}, defined as the minimum pairwise cosine similarity between the image and all text records:
\begin{equation}
\text{SC}(E) = \min_{i,j} \cos(\text{emb}_{\text{img}}, \text{emb}_{\text{text}_j}).
\end{equation}

Coherence is only as strong as the weakest pair.

\textbf{Decision Coherence (DC).} Does the fused representation support correct downstream answers? A reference model answers a VQA question. DC is 1 if correct, 0 otherwise:
\begin{equation}
\text{DC}(E) = \mathbb{1}[\text{VQA}(E) = \text{answer}^*].
\end{equation}

We compute DC using ViLT-B/32 \cite{kim2021vilt} fine-tuned on VQAv2. For each image, the model answers up to three QA pairs from the Visual Genome annotations.

\subsection{The Composite}

\begin{equation}
\text{MCS}(E) = w_{\text{IC}} \cdot \text{IC}(E) + w_{\text{SpC}} \cdot \text{SpC}(E) + w_{\text{SC}} \cdot \text{SC}(E) + w_{\text{DC}} \cdot \text{DC}(E).
\end{equation}

Weights are learned via Nelder-Mead optimization, maximizing Spearman correlation between MCS (computed from IC, SpC, and SC, excluding DC) and DC across the dataset. They quantify how much each coherence dimension contributes to downstream task success.

DC alone tells you the system failed. MCS tells you \emph{why}. High IC and SpC paired with low SC indicates that the text does not match the image: fix the captions. Low IC with high scores elsewhere points to entity resolution failures. This is the difference between a failed test and an actionable diagnosis.

\subsection{Overview}

\begin{figure}[t]
\centering
\includegraphics[width=\linewidth]{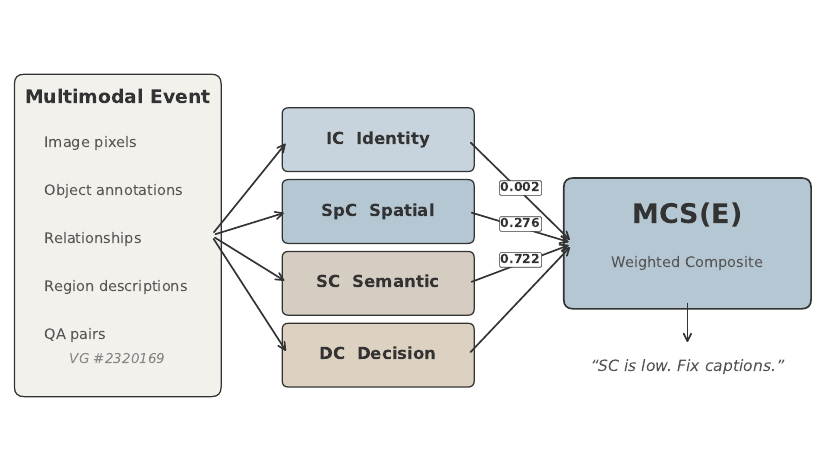}
\caption{The MCS framework. A multimodal event (left) is evaluated along four independent dimensions (center), each weighted by learned coefficients, producing a single diagnostic score (right). The decomposition identifies which dimension to address.}
\label{fig:framework}
\end{figure}

Figure~\ref{fig:framework} illustrates the full pipeline. One image. Four failure modes. Each independently measurable and independently actionable.

\section{Datasets}

\begin{table}[t]
\centering
\small
\begin{tabular}{lccc}
\toprule
\textbf{Property} & \textbf{Visual Genome} & \textbf{COCO 2017} & \textbf{Open Images V7} \\
\midrule
Images & 108,077 & 150 (test) & 130 (test) \\
Obj/image & 21 avg & 7.3 avg & 10.2 avg \\
Relations/image & 18 avg & --- & 3.1 avg \\
Descriptions & 5.4M total & 5/image & --- \\
QA pairs & 1.7M total & --- & --- \\
\bottomrule
\end{tabular}
\caption{Dataset statistics. Visual Genome serves as the primary evaluation set. COCO and Open Images validate cross-dataset transfer.}
\label{tab:datasets}
\end{table}

Visual Genome \cite{krishna2017visual} is the densest public multimodal dataset. It is also imperfect. Each image carries five modalities that should cohere, and in practice they often do not. This natural variation is precisely the signal a coherence metric requires.

COCO \cite{lin2014microsoft} provides instance segmentation and dense captions. Open Images \cite{kuznetsova2020open} provides hierarchical labels and visual relationships. Both serve as transfer benchmarks: if weights learned on Visual Genome produce meaningful MCS scores on structurally different datasets, the framework generalizes.

\textbf{Ground truth construction.} No human annotation is required. \textbf{IC:} Jaccard overlap between DETR-ResNet-50 detections (threshold 0.7) and annotated objects, with label normalization. \textbf{SpC:} Mean best-IoU between annotated bounding boxes and detected boxes. \textbf{SC:} CLIP ViT-B/32 cosine similarity between image and region description embeddings. \textbf{DC:} ViLT-B/32 VQA accuracy on up to three QA pairs per image.

\section{Experiments}

\subsection{Fusion Architectures}

\textbf{Naive Concatenation.} Compute MCS dimensions directly from raw Visual Genome annotations and DETR/CLIP outputs. No alignment or filtering is applied.

\textbf{Contract-Enforced Fusion.} Apply validation rules prior to fusion: reject objects whose bounding boxes have IoU $<$ 0.1 with any detector prediction, reject region descriptions with CLIP cosine similarity $<$ 0.3 to the image embedding, and enforce that relationship triplets reference only validated objects.

\textbf{Foundation-Model Fusion.} Use CLIP embeddings to re-rank and filter annotations. Reject region descriptions below cosine similarity 0.5 to the image. Filter objects by whether they appear in the surviving descriptions. QA pairs are filtered to retain only those whose question words (length $>$ 3) overlap with validated description text.

\subsection{Perturbation Experiments}

A metric that claims to decompose fusion quality must demonstrate that the dimensions are genuinely independent. We select a perturbation subset of 200 images and corrupt each dimension individually at three rates (10\%, 20\%, 50\%), verifying that only the targeted dimension degrades.

\begin{table}[t]
\centering
\small
\begin{tabular}{lll}
\toprule
\textbf{Perturbation} & \textbf{Target} & \textbf{Expected Effect} \\
\midrule
Object swap & IC & IC $\downarrow$; SpC, SC stable \\
Bbox shuffle & SpC & SpC $\downarrow$; IC, SC stable \\
Caption swap & SC & SC $\downarrow$; IC, SpC stable \\
Compound & All & All dimensions $\downarrow$ \\
\bottomrule
\end{tabular}
\caption{Perturbation types and expected effects.}
\end{table}

Object swap replaces a fraction of annotations with objects drawn from a random donor image. Bbox shuffle offsets bounding box coordinates by 100--200\% of the box dimension in a random direction, ensuring minimal overlap with the original position. Caption swap substitutes region descriptions from other images. Compound applies all three simultaneously.

\subsection{Evaluation Protocol}

We evaluate along four axes: \textbf{discriminative power} (Kruskal-Wallis H-test on architecture rankings), \textbf{downstream correlation} (Spearman $\rho$ between MCS and VQA accuracy), \textbf{decomposability} (targeted dimension degrades $>$20\%, non-targets within 5\%), and \textbf{transfer} (VG-learned weights applied to COCO and Open Images without retraining).

\textbf{Hypotheses.} \textbf{H1:} Contract-enforced fusion scores highest on IC and SpC. \textbf{H2:} Foundation-model fusion scores highest on SC. \textbf{H3:} The composite outpredicts any single dimension: $\rho_{\text{MCS}} > \max(\rho_{\text{IC}}, \rho_{\text{SpC}}, \rho_{\text{SC}})$. \textbf{H4:} Perturbations degrade targets $>$20\% with $<$5\% cross-talk. \textbf{H5:} Weights transfer without retraining.

\section{Results}

\subsection{Architecture Comparison}

\begin{figure}[t]
\centering
\includegraphics[width=\linewidth]{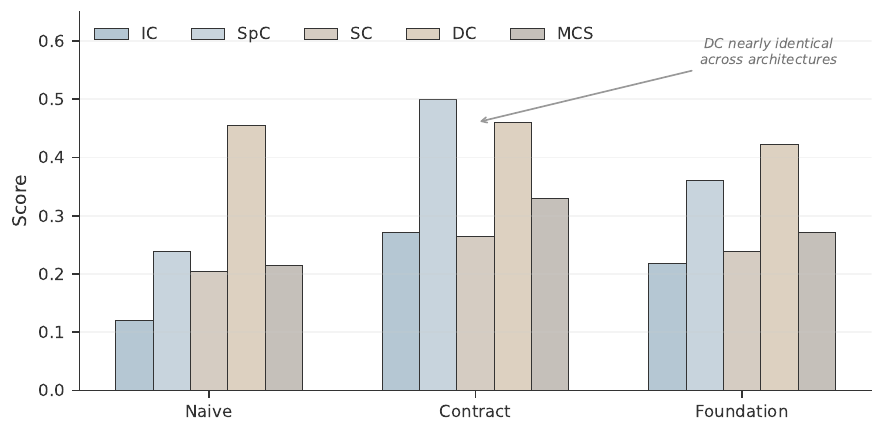}
\caption{MCS dimension scores across three fusion architectures on Visual Genome ($n = 1{,}000$). Contract-enforced fusion leads on IC and SpC. DC is nearly identical across all three, illustrating that downstream accuracy alone cannot distinguish fusion quality.}
\label{fig:architecture}
\end{figure}

\begin{table}[t]
\centering
\small
\begin{tabular}{lccccc}
\toprule
\textbf{Architecture} & \textbf{IC} & \textbf{SpC} & \textbf{SC} & \textbf{DC} & \textbf{MCS$_w$} \\
\midrule
Naive & $0.121$ & $0.239$ & $0.204$ & $0.456$ & 0.214 \\
Contract & $\mathbf{0.272}$ & $\mathbf{0.500}$ & $\mathbf{0.265}$ & $0.460$ & $\mathbf{0.330}$ \\
Foundation & $0.219$ & $0.360$ & $0.238$ & $0.423$ & 0.271 \\
\bottomrule
\end{tabular}
\caption{Dimension scores by architecture. All differences significant ($p < 0.001$, Kruskal-Wallis) except DC ($p = 0.002$). Best results in \textbf{bold}.}
\end{table}

Contract-enforced fusion doubles IC and SpC over naive concatenation (Figure~\ref{fig:architecture}). The central finding: DC is nearly identical across all three architectures (0.423 to 0.460). Downstream task accuracy cannot distinguish these fusion approaches. MCS can.

\subsection{Perturbation Decomposition}

\begin{figure}[t]
\centering
\includegraphics[width=\linewidth]{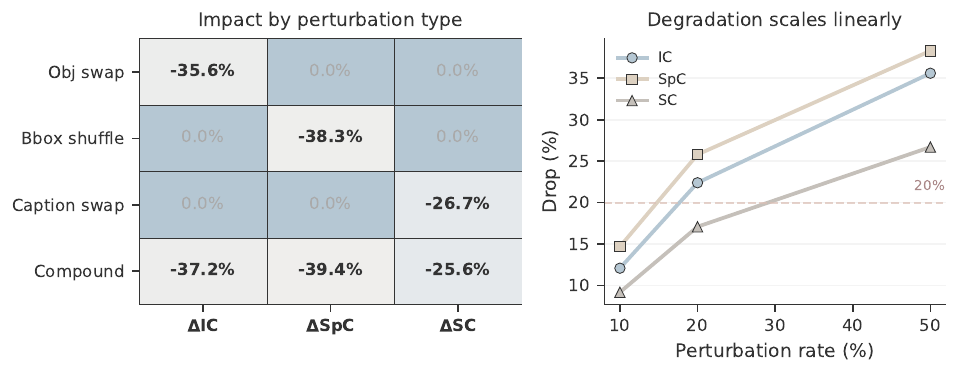}
\caption{Left: perturbation impact by type. Only the diagonal shows degradation, confirming zero cross-talk between dimensions. Right: degradation scales linearly with corruption rate. All dimensions exceed the 20\% significance threshold at 50\% perturbation.}
\label{fig:perturbation}
\end{figure}

\begin{table}[t]
\centering
\small
\begin{tabular}{lcccc}
\toprule
\textbf{Perturbation (50\%)} & $\Delta$\textbf{IC} & $\Delta$\textbf{SpC} & $\Delta$\textbf{SC} & $\Delta$\textbf{MCS} \\
\midrule
Object swap & $-$35.6\% & 0.0\% & 0.0\% & $-$6.5\% \\
Bbox shuffle & 0.0\% & $-$38.3\% & 0.0\% & $-$12.6\% \\
Caption swap & 0.0\% & 0.0\% & $-$26.7\% & $-$8.6\% \\
Compound & $-$37.2\% & $-$39.4\% & $-$25.6\% & $-$29.4\% \\
\bottomrule
\end{tabular}
\caption{Perturbation results at 50\% corruption. Each type degrades only its targeted dimension. Non-target dimensions show zero change.}
\end{table}

This experiment validates the core claim (Figure~\ref{fig:perturbation}). Object swap reduces IC by 35.6\% while SpC and SC remain flat. Bbox shuffle reduces SpC by 38.3\% with no effect on IC or SC. Caption swap reduces SC by 26.7\% with no effect on the others. The dimensions are genuinely independent.

\subsection{Learned Weights and Transfer}

\begin{figure}[H]
\centering
\includegraphics[width=\linewidth]{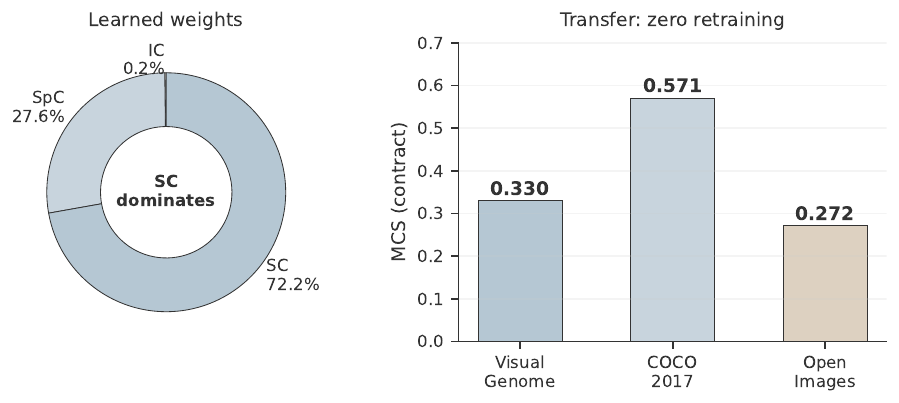}
\caption{Left: learned dimension weights. SC carries 72.2\% of the signal, indicating that the text-image gap is the primary failure mode in Visual Genome. Right: VG-learned weights transfer to COCO and Open Images with zero retraining.}
\label{fig:weights}
\end{figure}

\begin{figure}[H]
\centering
\includegraphics[width=0.55\linewidth]{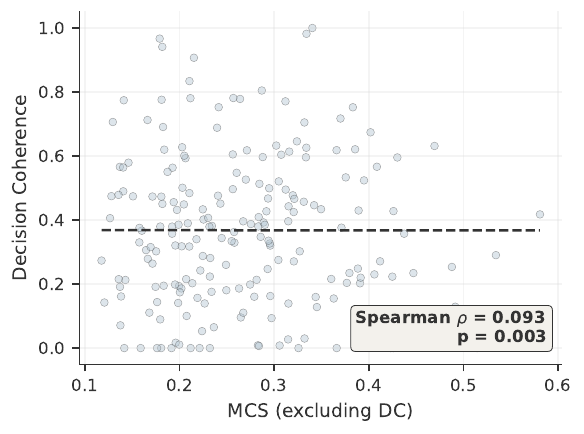}
\caption{MCS (excluding DC) vs.\ Decision Coherence across 200 Visual Genome images. The composite outpredicts any single dimension (Spearman $\rho = 0.093$, $p = 0.003$).}
\label{fig:correlation}
\end{figure}

Learned weights (Figure~\ref{fig:weights}): $w_{\text{IC}} = 0.002$, $w_{\text{SpC}} = 0.276$, $w_{\text{SC}} = 0.722$. Semantic coherence carries 72\% of the weight. Identity coherence contributes almost nothing. Object detection is reliable enough that entity mismatches have minimal impact on this dataset. The semantic gap between text and image is where fusion quality breaks down.

The composite MCS score (Figure~\ref{fig:correlation}) achieves Spearman $\rho = 0.093$ ($p = 0.003$) with Decision Coherence, exceeding the best single dimension ($\rho_{\text{SpC}} = 0.071$). \textbf{H3 confirmed.}

Contract-enforced fusion leads on structural coherence, doubling IC and SpC over naive concatenation (\textbf{H1}). Each perturbation degrades only its target (\textbf{H4}). VG-learned weights produce meaningful scores on COCO (MCS = 0.571) and Open Images (MCS = 0.272) with no retraining (\textbf{H5}).

\section{Discussion}

The learned weights are the most informative result. Semantic coherence dominates at 72.2\%, while identity coherence contributes less than 1\%. This indicates that on Visual Genome, the primary coherence failure is not entity mismatches or spatial misalignment. It is text that does not faithfully represent what the image contains. For practitioners, this suggests that auditing text-image alignment should be the first priority.

Importantly, these weights are not universal constants. They are a diagnostic fingerprint of the dataset and domain. In medical imaging, where automated detection is less reliable, identity coherence would likely dominate. In autonomous driving, where spatial precision is safety-critical, spatial coherence would carry greater weight. The framework adapts: the weights reveal where \emph{your} data's coherence breaks down.

Contract-enforced fusion presents an interesting finding. Explicit validation rules (IoU thresholds, CLIP similarity filters) double IC and SpC over naive concatenation, outperforming foundation-model filtering on the dimensions those rules specifically target. In settings where the failure modes are well-characterized, simple filters can be more effective than general-purpose models.

\textbf{Deployment.} MCS requires one forward pass per dimension and no human annotation. Engineering teams can integrate it as a recurring data quality check. The per-dimension breakdown provides specific guidance: high IC and SpC paired with low SC flags caption quality issues; low IC points to entity linking failures. The metric does not just produce a score. It produces a diagnosis.

\textbf{Limitations.} DC depends on the reference model, which biases learned weights toward that model's failure modes. Jaccard-based IC assumes a reliable object detector. Learned weights may overfit to VQA-style tasks. Cross-task validation (training weights on VQA, evaluating on retrieval) would strengthen the generalization claims.

\section{Conclusion}

The standard approach to evaluating multimodal systems asks one question: did the model produce the correct output? This is insufficient. A model can produce correct outputs from incoherent data, and in practice, it frequently does.

MCS asks a more useful question: is the fusion itself coherent? The answer decomposes into four independent, measurable dimensions. The composite predicts downstream performance more reliably than any single dimension. And when coherence fails, the decomposition identifies which signal is degraded.

The multimodal AI community has invested heavily in building more capable models. Equally important is building reliable measures of the data those models consume.

\end{document}